\documentclass[11pt]{article}

\usepackage[final]{acl}

\usepackage{times}
\usepackage{latexsym}

\usepackage[T1]{fontenc}

\usepackage[utf8]{inputenc}

\usepackage{microtype}

\usepackage{inconsolata}

\usepackage{graphicx}

\usepackage{amsmath,amssymb,amsthm}
\usepackage{booktabs}
\usepackage{multirow}
\usepackage{algorithm}
\usepackage{algpseudocode}

\usepackage{enumitem}
\usepackage{float}
\newtheorem{proposition}{Proposition}

\newtheorem{remark}{Remark}
\newtheorem{lemma}{Lemma}

\usepackage{xcolor}
\usepackage{colortbl}

\hypersetup{colorlinks=true,
  linkcolor=[HTML]{000099},
  citecolor=[HTML]{000099},
  urlcolor=[HTML]{000099}}

\title{SARATHI: Decoupled Data-Free Probing and Adaptive \\
 Reconstruction for Structured LLM Pruning}

\author{%
  \textbf{Sankar Behera}\textsuperscript{1},
  \textbf{Dhruv Singh}\textsuperscript{2},
  \textbf{Satyadev Ahlawat}\textsuperscript{3},
  \textbf{Yamuna Prasad}\textsuperscript{1}\\
  \textsuperscript{1}Department of Computer Science and Engineering, IIT Jammu\\
  \textsuperscript{2}Department of Mathematics, IIT Jammu\\
  \textsuperscript{3}Department of Electrical Engineering, IIT Jammu\\
  \texttt{\{sankar.behera, 2023uma0210, satyadev.ahlawat, yamuna.prasad\}@iitjammu.ac.in}
}

\begin{document}
\maketitle

\begin{abstract}
Structured pruning reduces the inference cost of large language models (LLMs) by permanently removing neurons and weight rows, yielding dense, hardware-agnostic speedup.
Existing post-training methods couple neuron selection and weight reconstruction through a shared calibration dataset, introducing domain bias and out-of-memory failures at high sparsity.
We propose SARATHI, a framework that decouples these two stages.
A data-free \emph{NMF Residual Subspace Probe} scores neurons using intrinsic weight geometry, without accessing any calibration data, and carries a formal connection to Robust PCA.
\emph{Adaptive OBS} then reconstructs the pruned weights with a localized calibration batch, operating in $\mathcal{O}(d^2)$ peak memory per layer and eliminating the global memory bottleneck of coupled methods.
A \emph{Bias Shift Compensation} step corrects systematic activation drift in post-LayerNorm architectures that would otherwise cause perplexity collapse at high sparsity.
Experiments across six models from two architecture families (gated and non-gated FFN) at 15--40\% sparsity show that SARATHI achieves competitive perplexity and matches or exceeds existing structured pruning baselines on zero-shot accuracy.
Code: \url{https://github.com/snkrbhr/SARATHI}
\end{abstract}

\section{Introduction}
\label{sec:intro}

LLMs deliver strong performance across downstream natural language tasks \citep{brown2020gpt3,touvron2023llama2,jiang2023mistral}, yet their billions of parameters impose prohibitive memory and latency costs at deployment. Structured pruning addresses this by physically removing entire rows or columns from weight matrices, yielding dense smaller matrices that accelerate inference on any BLAS hardware without requiring specialised sparse kernels, unlike unstructured methods \citep{frantar2023sparsegpt,sun2024wanda}. The proposed framework targets FFN intermediate neurons, where compression leverage is uniquely high: FFN sub-layers hold $2\times$ more parameters than attention in MHA models and $4.2$--$4.7\times$ more in GQA-based models; and non-linear activations (ReLU, SwiGLU) are exclusive to FFN layers, so pruning simultaneously reduces parameter count and activation compute \citep{mirzadeh2023relu}. Full parameter breakdowns are in Appendix~\ref{sec:appendix_ffn_motivation}.

\paragraph{The coupling flaw.}
Existing post-training structured pruning methods \citep{wei2024sobp,an2024flap,ashkboos2024slicegpt} share a fundamental design flaw: they use the \emph{same} calibration dataset $\mathcal{D}_\text{calib}$ to simultaneously (i) score which neurons to prune and (ii) compute the weight-compensation matrix. This coupling is computationally costly: building the activation Hessian $\mathbf{H} = \mathbf{X}_\text{calib}\mathbf{X}_\text{calib}^\top \in \mathbb{R}^{d \times d}$ per layer requires $\mathcal{O}(d^2)$ working memory per layer; in practice, frameworks accumulate calibration activations, inverse Hessians, Cholesky factors, and OBS buffers globally, incurring substantial peak GPU memory at scale. This is statistically harmful, biasing the pruning mask toward neurons active on $\mathcal{D}_\text{calib}$ and away from neurons important for open-domain generalization. As demonstrated in \S\ref{sec:experiments}, this causes OOM failures at high sparsity.

\paragraph{SARATHI.}
The proposed framework decouples the problem into two independent subproblems (\S\ref{sec:decouple}): (P1) a \emph{data-free} selection stage using the intrinsic geometry of the weight matrix, and (P2) a localized reconstruction stage with $\mathcal{O}(d^2)$ peak memory per layer. The selection probe (\S\ref{sec:nmf}) applies NMF to $|\mathbf{W}^{(\ell)}|$, identifying neurons whose weight rows lie outside the shared low-rank non-negative subspace as structurally irreplaceable (a formal connection to Robust PCA \citep{candes2011rpca}), which is novel in the post-training LLM structured pruning setting. The framework also identifies and corrects a \emph{cascading bias shift} in post-LayerNorm architectures (\S\ref{sec:obs}), reducing WikiText-2 PPL on OPT-2.7B at 40\% from 2,082 to 18.98 (99.1\% reduction vs.\ the uncorrected model).

\paragraph{Contributions.}
\begin{itemize}[leftmargin=*,nosep]
  \item \textbf{Decoupled Probe-Reconstruct} architecture eliminating calibration bias and the $\mathcal{O}(Ld^2)$ memory bottleneck (\S\ref{sec:decouple}).
  \item \textbf{NMF Residual Subspace Probing}: data-free neuron scoring with a Robust PCA connection and formal pruning-safety bound (Proposition~\ref{prop:nmf}, \S\ref{sec:nmf}).
  \item \textbf{Bias Shift Compensation}: zero-overhead closed-form correction for post-LayerNorm PPL collapse (\S\ref{sec:obs}).
\end{itemize}

Evaluation on six models (LLaMA-3-8B, Mistral-7B, OPT-2.7B/6.7B, LLaMA-2-7B/13B) at 15--40\% sparsity (\S\ref{sec:experiments}) shows competitive PPL and consistently higher or comparable zero-shot accuracy relative to state-of-the-art baselines across all architectures.

\section{The SARATHI Framework}
\label{sec:method}
\begin{figure*}[t]
  \centering
  \includegraphics[width=\textwidth]{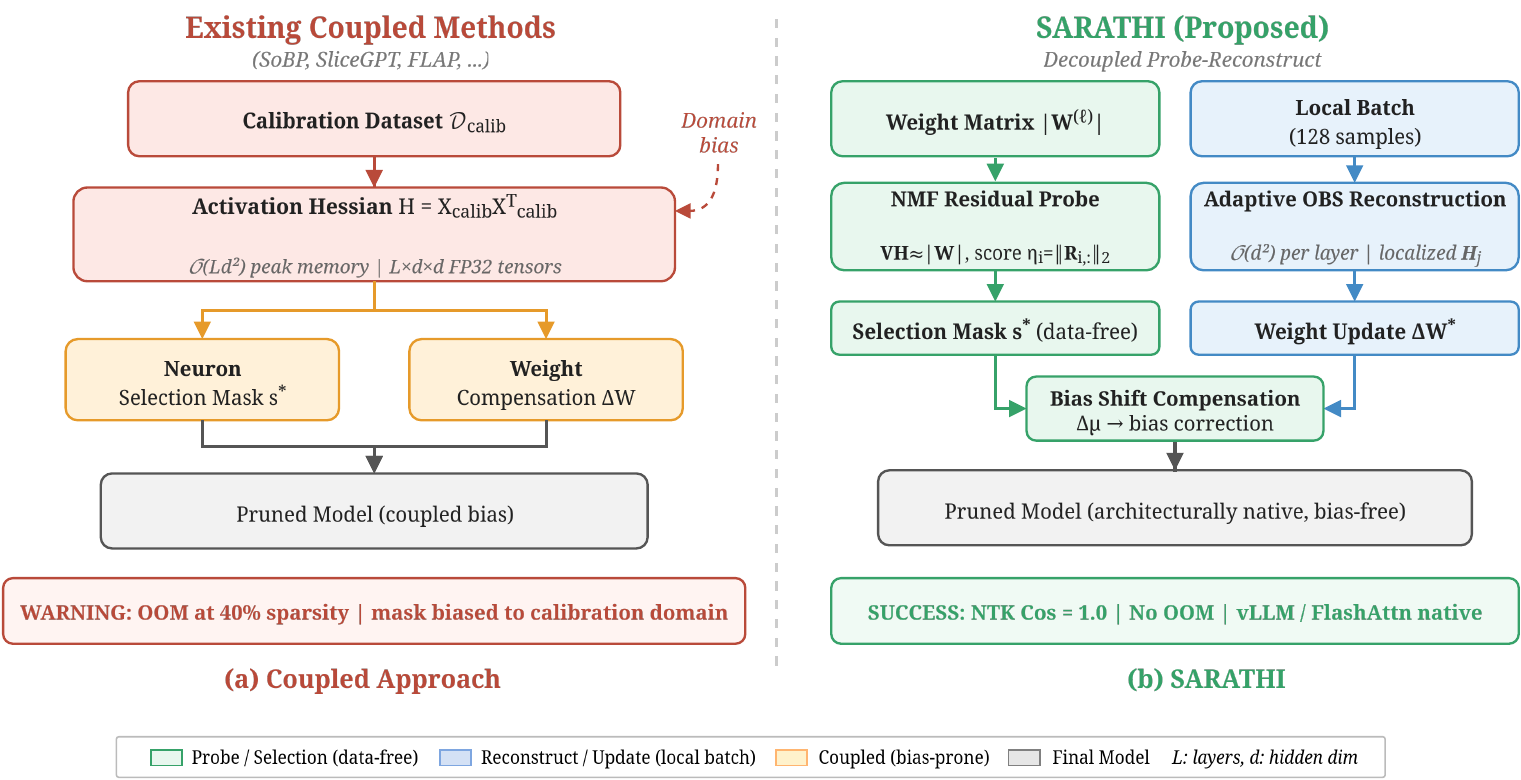}
  \caption{\textbf{The coupling flaw in existing structured pruning vs.\ SARATHI's decoupled design.}
    \emph{(a) Coupled methods} (SoBP, SliceGPT, FLAP) use the same calibration dataset $\mathcal{D}_\text{calib}$
    for both neuron selection and weight reconstruction, incurring $\mathcal{O}(Ld^2)$ peak memory and
    introducing calibration-domain bias into the pruning mask.
    \emph{(b) SARATHI} separates these two stages: neuron selection is entirely data-free
    (NMF Residual Probe, green), while weight reconstruction uses only a small localized batch
    with $\mathcal{O}(d^2)$ memory per layer (Adaptive OBS, blue).
    The result is an architecturally native pruned model with Neural Tangent Kernel (NTK) Cosine Similarity of 1.0000
    and no OOM failures, even at 40\% sparsity on OPT-6.7B.}
  \label{fig:architecture}
\end{figure*}

Figure~\ref{fig:architecture} illustrates the overall decoupled architecture.

\subsection{Problem Formulation and the Coupling Flaw}
\label{sec:coupling}

Consider a transformer with $L$ MLP layers. For the $\ell$-th layer, let $\mathbf{W}^{(\ell)} \in \mathbb{R}^{N \times d}$ be the projection weight matrix, where $N$ is the intermediate dimension and $d$ is the model hidden size. Given input activations $\mathbf{X} \in \mathbb{R}^{d \times T}$ over $T$ token positions, the layer output is $\mathbf{Z} = \mathbf{W}^{(\ell)}\mathbf{X} \in \mathbb{R}^{N \times T}$.

Structured pruning seeks a binary row-selection mask $\mathbf{s} \in \{0,1\}^N$ that retains exactly $\lfloor(1{-}\rho)N\rfloor$ rows (neurons) at target sparsity $\rho \in (0,1)$, yielding the compressed weight $\mathbf{W}^{(\ell)}_{\mathbf{s}} = \mathbf{W}^{(\ell)}[\mathbf{s},:]$. The ideal post-pruning weight correction $\Delta\mathbf{W}$ is co-optimized with $\mathbf{s}$, giving the \emph{coupled} objective solved by SoBP \citep{wei2024sobp}, SparseGPT \citep{frantar2023sparsegpt}, and SliceGPT \citep{ashkboos2024slicegpt}:
\begin{equation}
\begin{split}
\label{eq:coupled}
\min_{\mathbf{s},\,\Delta\mathbf{W}} &\left\| \mathbf{W}^{(\ell)}\mathbf{X}_\text{calib} - \bigl(\mathbf{W}^{(\ell)}_{\mathbf{s}} + \Delta\mathbf{W}\bigr)\mathbf{X}_\text{calib} \right\|_F^2 \\
&\text{s.t.} \quad \|\mathbf{s}\|_0 = \lfloor(1{-}\rho)N\rfloor
\end{split}
\end{equation}

\begin{remark}[NP-hardness and calibration coupling]
The $\|\mathbf{s}\|_0{=}K$ constraint is NP-hard \citep{natarajan1995sparse}. Fixing $\mathbf{X}_\text{calib}$ for both scoring and reconstruction introduces domain bias: the mask $\mathbf{s}^*_\text{calib}$ incurs positive excess risk on any $\mathcal{D} \not\approx \mathcal{D}_\text{calib}$.
\end{remark}

Beyond domain bias, computing the Hessian $\mathbf{H} = \mathbf{X}_\text{calib}\mathbf{X}_\text{calib}^\top \in \mathbb{R}^{d \times d}$ per layer requires $\mathcal{O}(d^2)$ working memory per layer. For OPT-6.7B with $L{=}32$, $d{=}4096$, the Hessians themselves occupy $\approx$2.1\,GB in FP32; in practice, peak GPU memory is substantially higher due to inverse Hessians, Cholesky factors, OBS reconstruction buffers, and calibration activation tensors held concurrently, causing OOM failures at 40\% sparsity as demonstrated empirically in \S\ref{sec:experiments}.

\subsection{Decoupled Probe-Reconstruct Architecture}
\label{sec:decouple}

The proposed method resolves both failure modes by decomposing Eq.~\ref{eq:coupled} into two \emph{sequentially independent} subproblems with disjoint information sources:
\noindent \textbf{(P1: Selection)}
\begin{equation}
\label{eq:probe}
\mathbf{s}^* = \operatorname*{arg\,max}_{\mathbf{s}:\|\mathbf{s}\|_0 = \lfloor(1-\rho)N\rfloor} \sum_{i=1}^N s_i \cdot \eta_i
\end{equation}
\noindent \textbf{(P2: Reconstruction)}
\begin{equation}
\begin{split}
\label{eq:reconstruct}
\Delta\mathbf{W}^* = \operatorname*{arg\,min}_{\Delta\mathbf{W}\in\mathcal{C}(\mathbf{s}^*)} &\left\| \bigl(\mathbf{W}^{(\ell)} - \Delta\mathbf{W}\bigr)\mathbf{X}_\text{local} \right. \\
&\left. - \mathbf{W}^{(\ell)}_{\mathbf{s}^*}\mathbf{X}_\text{local} \right\|_F^2
\end{split}
\end{equation}
where $\eta_i \in \mathbb{R}_{\geq 0}$ is a \emph{data-free} geometric importance score computed purely from $\mathbf{W}^{(\ell)}$ (derived in \S\ref{sec:nmf}), and $\mathcal{C}(\mathbf{s}^*)= \{\Delta\mathbf{W} : \Delta\mathbf{W}[i,{:}]{=}\mathbf{0}\;\forall\, s_i{=}1\}$ restricts compensation to only the pruned rows. The localized batch $\mathbf{X}_\text{local}$ (128 samples) is used \emph{only} in P2, requiring $\mathcal{O}(d^2)$ memory per layer rather than $\mathcal{O}(Ld^2)$.

\subsection{NMF Residual Subspace Probing}
\label{sec:nmf}

\paragraph{Low-rank structure of LLM weight matrices.}
Trained MLP weight matrices satisfy $\operatorname{rank}_\epsilon(\mathbf{W}^{(\ell)}) \ll \min(N,d)$ \citep{hu2022lora,svdllm2024}. The row space is well-approximated by a subspace $\mathcal{V}$ with $\dim(\mathcal{V}) = r \ll d$. Neurons with $\mathbf{W}^{(\ell)}_{i,:} \in \mathcal{V}$ can be removed with bounded error; neurons with $\mathbf{W}^{(\ell)}_{i,:} \notin \mathcal{V}$ are structurally irreplaceable.

\paragraph{Non-negative Matrix Factorization.}
To align with the non-negative activation regime of ReLU and SwiGLU FFN layers, our framework applies NMF to the element-wise absolute-value weight $|\mathbf{W}^{(\ell)}| \in \mathbb{R}^{N \times d}_{\geq 0}$, seeking factors $\mathbf{V} \in \mathbb{R}^{N \times r}_{\geq 0}$, $\mathbf{H} \in \mathbb{R}^{r \times d}_{\geq 0}$ that solve:
\begin{equation}
\label{eq:nmf}
\min_{\mathbf{V} \geq 0,\, \mathbf{H} \geq 0} \;\bigl\| |\mathbf{W}^{(\ell)}| - \mathbf{V}\mathbf{H} \bigr\|_F^2
\end{equation}
NMF is preferred over truncated Singular Value Decomposition (SVD) because its non-negativity constraint enforces an \emph{additive, parts-based} decomposition \citep{lee1999nmf}, aligning with the non-negative activation regime of ReLU/SwiGLU layers. Denoting the residual $\mathbf{S} = |\mathbf{W}^{(\ell)}| - \mathbf{VH}$, it reliably identifies structurally irreplaceable neurons. This choice is empirically validated in Appendix~\ref{sec:appendix_nmf_svd}: replacing NMF with a signed SVD residual raises PPL by more than 8{,}500 points on OPT-6.7B despite a 98.95\% overlap in top-ranked neurons, confirming that non-negativity in the factorization is the decisive factor. Updates converge to a stationary point:
\begin{equation}
\label{eq:nmf_update}
\begin{aligned}
\mathbf{V} &\leftarrow \mathbf{V} \odot \frac{|\mathbf{W}|\mathbf{H}^\top}{\mathbf{V}\mathbf{H}\mathbf{H}^\top + \epsilon}, \\
\mathbf{H} &\leftarrow \mathbf{H} \odot \frac{\mathbf{V}^\top|\mathbf{W}|}{\mathbf{V}^\top\mathbf{V}\mathbf{H} + \epsilon}
\end{aligned}
\end{equation}
We run these updates for 100 iterations (verified to converge within 50 iterations for all tested layers; Remark~\ref{rem:nmf}) with $\epsilon{=}10^{-8}$ for numerical stability.

\paragraph{NMF Residual Score.}
Given residual $\mathbf{S}$ and a threshold $\tau > 0$ (derived by global MAD thresholding in \S\ref{sec:mad}), define the element-wise binary mask $\mathcal{M}_{i,j} = \mathbb{I}(\mathbf{S}_{i,j} \geq \tau)$. The importance score of neuron $i$ is:
\begin{equation}
\label{eq:score}
\eta_i \;=\; \sum_{j} \bigl(|\mathbf{W}^{(\ell)}_{i,j}| \cdot \mathcal{M}_{i,j}\bigr)
\end{equation}
Neurons with $\eta_i = 0$ have no weight entries escaping the NMF subspace and are pruning-safe; neurons with $\eta_i > 0$ contain structurally irreplaceable weights outside the consensus manifold.

\begin{proposition}[Pruning safety bound for zero-$\eta$ neurons]
\label{prop:nmf}
Let $(\mathbf{V}^*, \mathbf{H}^*)$ be a stationary point of Eq.~\ref{eq:nmf}. For neuron $i$ with $\eta_i = 0$ (i.e.\ all weight entries of row $i$ are within the NMF subspace up to $\tau$), the output reconstruction error from removing row $i$ and applying Adaptive OBS compensation is bounded by $\mathcal{O}(\tau\sqrt{d}\|\mathbf{X}\|_F)$. Conversely, for neuron $j$ with $\eta_j > 0$, the row $\mathbf{W}^{(\ell)}_{j,:}$ contains entries outside $\mathrm{col}(\mathbf{V}^{*\top})$, making its removal irreversible within the low-rank subspace. Full derivation in Appendix~\ref{sec:proof_prop1}.
\end{proposition}

\paragraph{Connection to Robust PCA.}
The decomposition $|\mathbf{W}| = \mathbf{VH} + \mathbf{S}$ mirrors the Robust PCA decomposition $\mathbf{M} = \mathbf{L} + \mathbf{S}$ \citep{candes2011rpca}, where $\mathbf{L}$ is low-rank and $\mathbf{S}$ is sparse. High-$\eta_i$ rows of $\mathbf{S}$ correspond to the sparse outlier component: they encode information not captured by the consensus weight manifold and must be retained. This operationalises the principle of retaining the outlier subspace and pruning the low-rank component, providing the first formal geometric justification for a data-free structured pruning probe.

\begin{remark}[Stationarity and residual quality]
\label{rem:nmf}
Prop.~\ref{prop:nmf} holds at any stationary point \citep{lee1999nmf}. Empirically, per-layer Frobenius residual ratios for LLaMA-3-8B and OPT-6.7B are tightly concentrated in $[0.601, 0.614]$ across all 32 layers, confirming consistently informative stationary points. The error bound holds jointly with Adaptive OBS correction (Lemma~\ref{lem:obs}).
\end{remark}

\paragraph{SwiGLU gate-value coupling.}
For gated FFN architectures (e.g., LLaMA, Mistral), neuron $i$ coordinates across three matrices: gate ($\mathbf{W}^\text{gate}$), up ($\mathbf{W}^\text{up}$), and down ($\mathbf{W}^\text{down}$) projections \citep{dauphin2017glu}. To avoid the vanishing-score problem of multiplicative coupling (where a near-zero score in one projection nullifies the neuron), we use the arithmetic mean:
\begin{equation}
\label{eq:swiglu_couple}
\tilde{\eta}_i = \tfrac{1}{3}\bigl(\eta_i^\text{gate} + \eta_i^\text{up} + \eta_i^\text{down}\bigr)
\end{equation}
This ensures balanced structural retention across coupled matrices without disproportionately penalising asymmetric feature distributions.

\subsection{Robust Budget Allocation via MAD Thresholding}
\label{sec:mad}

Given per-layer score vectors $\boldsymbol{\eta}^{(\ell)} \in \mathbb{R}^{N_\ell}$, a global threshold $\tau$ is derived from the pooled scores. Mean/std-based thresholds are sensitive to the heavy-tailed distribution of $\eta_i$ in LLMs \citep{dettmers2022llmint8}, where a small fraction of neurons exhibit $\eta_i \gg \bar{\eta}$, inflating $\bar{\eta}$ and raising the threshold beyond the intended sparsity. Instead, we apply the \emph{Median Absolute Deviation} (MAD), with a breakdown point of 0.5 (the highest achievable for any location estimator):
\begin{equation}
\label{eq:mad}
\begin{aligned}
\tau &= \operatorname{median}(\boldsymbol{\eta}) - k \cdot \operatorname{MAD}(\boldsymbol{\eta}), \\
\operatorname{MAD}(\boldsymbol{\eta}) &= \operatorname{median}\!\left(|\boldsymbol{\eta} - \operatorname{median}(\boldsymbol{\eta})|\right)
\end{aligned}
\end{equation}
Neuron $i$ in layer $\ell$ is pruned iff $\eta_i^{(\ell)} < \tau$. The scalar $k > 0$ controls the global budget via binary search to match target $\rho$. An adaptive \emph{min-keep floor} $\alpha$ enforces $\|\mathbf{s}^{(\ell)}\|_0 \geq \lceil\alpha N_\ell\rceil$ per layer, preventing collapse, which is more robust than SoBP's per-architecture $\mu/\sigma$ normalisation \citep{wei2024sobp}.

\subsection{Adaptive OBS Weight Reconstruction}
\label{sec:obs}

Unlike standard OBS frameworks \citep{hassibi1993obs, frantar2023sparsegpt} that jointly use the inverse Hessian for pruning saliency and weight updates, SARATHI decouples structural search from compensation. By pre-determining the global mask $\mathbf{s}^*$ via the NMF probe, we bypass layer-wise saliency evaluation. This transforms OBS from a joint search-and-update problem into a fixed-constraint optimization: the Hessian strictly computes compensatory updates for the pre-determined survivors. This decoupling enables asymmetrical dimension allocation across layers while preserving local error-compensation guarantees, yielding a closed-form solution for Subproblem P2 (Eq.~\ref{eq:reconstruct}):

\begin{lemma}[Column-wise OBS update]
\label{lem:obs}
Define the regularized empirical Hessian $\mathbf{H}^{\text{obs}}_j = \mathbf{X}_\text{local}\mathbf{X}_\text{local}^\top + \lambda\mathbf{I} \in \mathbb{R}^{d \times d}$ for column index $j$ (distinct from the NMF factor $\mathbf{H}$). The minimum-perturbation weight update that removes the contribution of pruned row $i$ at column $j$, derived by zeroing the $(i,j)$ entry subject to minimizing $\frac{1}{2}\delta\mathbf{w}^\top \mathbf{H}^{\text{obs}}_j\, \delta\mathbf{w}$ via Lagrange conditions, is:
\begin{equation}
\label{eq:obs}
\delta w_{ij} = -\frac{w_{ij}}{[{\mathbf{H}^{\text{obs}}_j}^{-1}]_{jj}} \cdot [{\mathbf{H}^{\text{obs}}_j}^{-1}]_{j,:}
\end{equation}
with induced saliency (pruning cost) $\mathcal{S}_{ij} = \frac{w_{ij}^2}{2[{\mathbf{H}^{\text{obs}}_j}^{-1}]_{jj}}$.
\end{lemma}

This update is applied iteratively over all pruned row indices for each column $j \in \{1,\ldots,d\}$ of the surviving weight matrix. The damping factor $\lambda > 0$ ensures ${\mathbf{H}^{\text{obs}}_j}^{-1}$ is well-conditioned. Because $\mathbf{H}^{\text{obs}}_j$ is constructed from only 128 local samples, peak memory consumption is $\mathcal{O}(d^2)$ per layer, a factor of $L$ reduction relative to the global coupled Hessian.

\begin{figure}[t]
  \centering
  \includegraphics[width=\columnwidth]{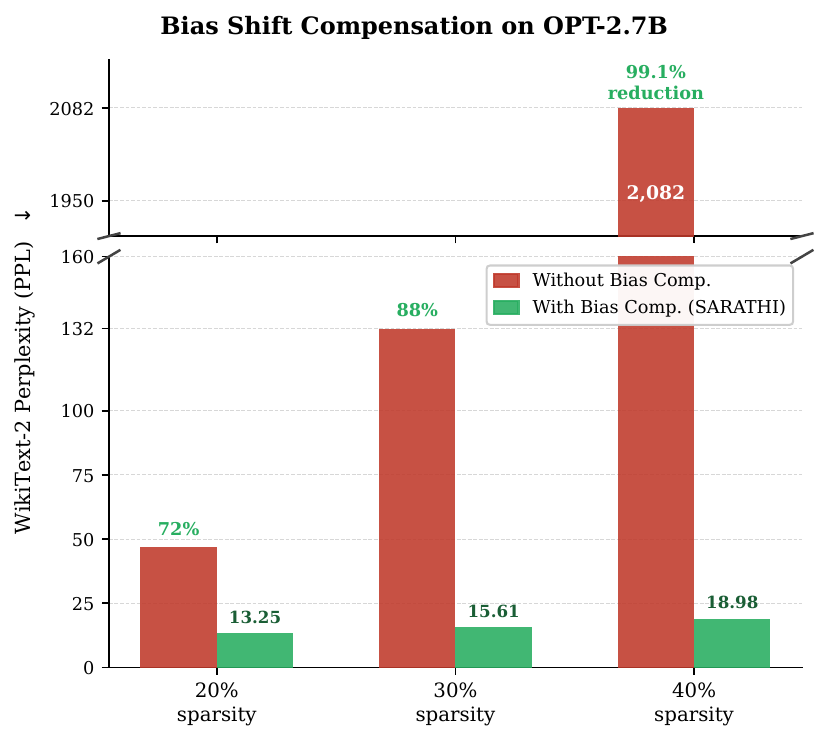}
  \caption{\textbf{Bias Shift Compensation on OPT-2.7B PPL.}
    Without correction (red), PPL collapses to 2,082 at 40\% sparsity.
    SARATHI's zero-overhead bias update (Eq.~\ref{eq:bias}, green) reduces PPL by
    72\%, 88\%, and \textbf{99.1\%} at 20\%, 30\%, and 40\%, respectively.}
  \label{fig:bias_comp}
\end{figure}

\paragraph{Bias Shift Compensation for post-LayerNorm architectures.}
In models using post-LayerNorm (e.g., the OPT family \citep{zhang2022opt}), the LayerNorm at layer $\ell{+}1$ operates on $\mathbf{z}^{(\ell)} = f(\mathbf{W}^{(\ell)}\mathbf{x}) \in \mathbb{R}^d$. Removing $K$ neurons shifts the empirical mean of $\mathbf{z}^{(\ell)}$ by:
\begin{equation}
\begin{split}
\label{eq:bias_shift}
\Delta\boldsymbol{\mu}^{(\ell)} &= \mathbb{E}_{\mathbf{x}\sim\mathbf{X}_\text{local}}\!\left[f\!\left(\mathbf{W}^{(\ell)}\mathbf{x}\right)\right] \\
&\quad - \mathbb{E}_{\mathbf{x}\sim\mathbf{X}_\text{local}}\!\left[f\!\left(\mathbf{W}^{(\ell)}_{\mathbf{s}^*}\mathbf{x}\right)\right] \;\in\; \mathbb{R}^{d}
\end{split}
\end{equation}
This shift propagates through the LayerNorm scale-and-shift parameters, cascading across all $L{-}\ell$ subsequent layers and producing $\mathcal{O}(K \cdot (L-\ell))$ cumulative activation error. We correct this analytically with a zero-overhead bias update:
\begin{equation}
\label{eq:bias}
\hat{\mathbf{b}}^{(\ell)} \;=\; \mathbf{b}^{(\ell)} + \Delta\boldsymbol{\mu}^{(\ell)}
\end{equation}
computed from the same $\mathbf{X}_\text{local}$ used in Adaptive OBS with no additional forward passes. The empirical effect is visualised in Figure~\ref{fig:bias_comp}: PPL on OPT-2.7B at 40\% sparsity drops from 2,082 to 18.98 upon applying Eq.~\ref{eq:bias}.

\begin{algorithm}[t]
\caption{SARATHI Pruning Pipeline}
\label{alg:sarathi}
\small
\begin{algorithmic}[1]
\Require Model $\mathcal{M}$, target sparsity $\rho$, NMF rank $r$, min-keep $\alpha$
\Ensure Pruned model $\mathcal{M}^*$
\For{each MLP layer $\ell$}
    \State Compute $|\mathbf{W}^{(\ell)}|$; solve NMF: $|\mathbf{W}^{(\ell)}| \approx \mathbf{V}\mathbf{H}$ (Eq.~\ref{eq:nmf})
    \State Compute residual $\mathbf{S} = |\mathbf{W}^{(\ell)}| - \mathbf{V}\mathbf{H}$; set $\mathcal{M}_{i,j} = \mathbb{I}(\mathbf{S}_{i,j}\!\geq\!\tau)$
    \State Score neurons: $\eta_i = \sum_{j}|\mathbf{W}^{(\ell)}_{i,j}|\cdot\mathcal{M}_{i,j}$ (Eq.~\ref{eq:score})
    \If{SwiGLU architecture}
        \State Couple gate-value scores: $\tilde{\eta}_i = \tfrac{1}{3}(\eta_i^\text{gate} + \eta_i^\text{up} + \eta_i^\text{down})$
    \EndIf
\EndFor
\State Apply global MAD threshold to scores (Eq.~\ref{eq:mad}); respect min-keep $\alpha$
\For{each MLP layer $\ell$}
    \State Remove neurons where $\eta_i < \tau$; compute $\Delta\mu^\ell$
    \State Run Adaptive OBS reconstruction (Eq.~\ref{eq:obs})
    \State Apply bias shift compensation: $\mathbf{b}^\ell \mathrel{+}= \Delta\mu^\ell$ (Eq.~\ref{eq:bias})
    \State Physically resize $\mathbf{W}^{(\ell)}_\text{down}$, $\mathbf{W}^{(\ell)}_\text{up}$, $\mathbf{W}^{(\ell)}_\text{gate}$
\EndFor
\State \Return pruned $\mathcal{M}^*$
\end{algorithmic}
\end{algorithm}

\section{Experiments}
\label{sec:experiments}

\subsection{Experimental Setup}
\label{sec:setup}

\paragraph{Models.} Experiments span two architectural families: \textit{gated FFN} (LLaMA-3-8B \citep{metallamathree2024}, Mistral-7B \citep{jiang2023mistral}, LLaMA-2-7B/13B \citep{touvron2023llama2}) and \textit{non-gated FFN} (OPT-2.7B, OPT-6.7B \citep{zhang2022opt}), covering 2.7B--13B parameters.

\paragraph{Baselines.} We compare against SoBP \citep{wei2024sobp}, SliceGPT \citep{ashkboos2024slicegpt}, FLAP \citep{an2024flap}, Dynamic Slicing \citep{dumitru2024change}, and PruneNet \citep{prunenet2025}. Two internal variants isolate the probe contribution: SARATHI-NMF (data-free NMF residual) and SARATHI-Wanda (activation-based).

\paragraph{Metrics.} WikiText-2 PPL ($\downarrow$, seq.\ len.\ 2048); average zero-shot accuracy ($\uparrow$) on 7 commonsense tasks via lm-eval-harness \citep{gao2021lmeval}; inference throughput (tokens/sec) on NVIDIA A100 at BS=1 and BS=16.

\paragraph{Implementation.} NMF rank $r{=}16$, min-keep floor $\alpha{=}0.5$ (sensitivity in Table~\ref{tab:ablation}); Adaptive OBS reconstruction uses 128 calibration samples (C4 \citep{raffel2020t5} for OPT, WikiText-2 \citep{merity2017wikitext} default), $\lambda{=}0.01$; single A100 40GB GPU, no retraining.

\subsection{Main Results}
\label{sec:main_results}

\begin{figure}[!htbp]
  \centering
  \includegraphics[width=\columnwidth]{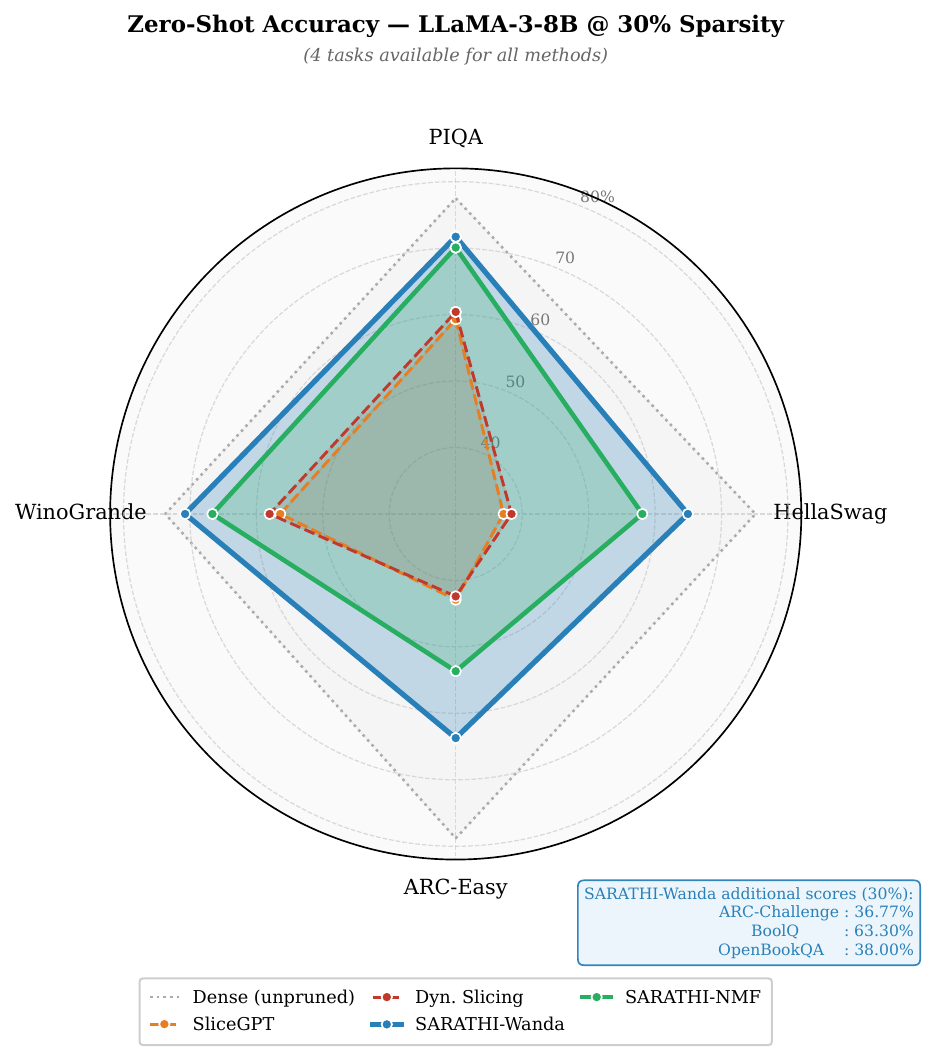}
  \caption{\textbf{Zero-shot accuracy for LLaMA-3-8B at 30\% sparsity.}
    SARATHI-Wanda (blue) and SARATHI-NMF (green) dominate SliceGPT (orange) and Dynamic Slicing (red).
    Grey dotted polygon = dense baseline.
    SliceGPT/Dyn.~Slicing reported only on 4 tasks \citep{dumitru2024change}.}
  \label{fig:radar}
\end{figure}

Table~\ref{tab:main_ppl} reports WikiText-2 PPL across all models and sparsity levels.
Table~\ref{tab:main_zs} reports zero-shot accuracy at 30\% sparsity. SoBP results cover OPT and LLaMA-2-13B models \citep{wei2024sobp}; SliceGPT and Dynamic Slicing results for LLaMA-3-8B and Mistral-7B are taken from \citet{dumitru2024change}.

\begin{table}[t]
\caption{WikiText-2 perplexity ($\downarrow$) across 15--40\% sparsity. \textbf{Bold} indicates the best result. OOM = out-of-memory failure. $\dagger$ indicates failure due to code errors. $\ddagger$ indicates results not reported by the original authors. SliceGPT and Dynamic Slicing results for LLaMA-3-8B and Mistral-7B are from \citet{dumitru2024change}. SliceGPT OPT baselines are from \citet{wei2024sobp}. All SARATHI models are compressed without retraining.}
\label{tab:main_ppl}
\centering
\small
\setlength{\tabcolsep}{4pt}
\resizebox{\columnwidth}{!}{%
\begin{tabular}{lcccc}
\toprule
\textbf{Method} & \textbf{15\%} & \textbf{20\%} & \textbf{30\%} & \textbf{40\%} \\
\midrule
\multicolumn{5}{l}{\textit{LLaMA-3-8B} (Dense PPL: 6.14)} \\
SliceGPT \citep{ashkboos2024slicegpt} & $\ddagger$ & $\ddagger$ & 13.37 & 20.69 \\
Dyn.~Slicing \citep{dumitru2024change} & $\ddagger$ & $\ddagger$ & 12.96 & 19.11 \\
\rowcolor[gray]{0.93}SARATHI-Wanda  & 7.78  & 8.18  & 9.04 & 10.10 \\
\rowcolor[gray]{0.93}SARATHI-NMF    & \textbf{7.28} & \textbf{7.70} & \textbf{8.70} & \textbf{9.92} \\
\midrule
\multicolumn{5}{l}{\textit{Mistral-7B} (Dense PPL: 5.32)} \\
SliceGPT \citep{ashkboos2024slicegpt} & $\ddagger$ & $\ddagger$ & 8.87  & 13.33 \\
Dyn.~Slicing \citep{dumitru2024change} & $\ddagger$ & $\ddagger$ & 8.76  & 12.62 \\
\rowcolor[gray]{0.93}SARATHI-Wanda  & 6.74  & 7.04  & 7.74 & 8.59 \\
\rowcolor[gray]{0.93}SARATHI-NMF    & \textbf{6.45} & \textbf{6.71} & \textbf{7.41} & \textbf{8.38} \\
\midrule
\multicolumn{5}{l}{\textit{OPT-2.7B} (Dense PPL: 12.47)} \\
FLAP \citep{an2024flap} & --    & 15.38 & 18.32 & $\dagger$ \\
SliceGPT \citep{ashkboos2024slicegpt} & 13.28 & 13.89 & 16.31 & $\dagger$ \\
SoBP \citep{wei2024sobp} & 12.55 & \textbf{12.72} & \textbf{13.89} & $\dagger$ \\
\rowcolor[gray]{0.93}SARATHI-Wanda  & 12.74 & 13.48 & 15.89 & \textbf{18.72} \\
\rowcolor[gray]{0.93}SARATHI-NMF    & \textbf{12.52} & 13.25 & 15.61 & 18.98 \\
\midrule
\multicolumn{5}{l}{\textit{OPT-6.7B} (Dense PPL: 10.86)} \\
FLAP \citep{an2024flap} & --    & 12.80 & 15.42 & $\dagger$ \\
SliceGPT \citep{ashkboos2024slicegpt} & 11.26 & 11.60 & 12.80 & $\dagger$ \\
SoBP \citep{wei2024sobp} & \textbf{11.02} & \textbf{11.32} & \textbf{12.17} & $\dagger$ \\
\rowcolor[gray]{0.93}SARATHI-Wanda  & 11.35 & 12.06 & 14.02 & OOM \\
\rowcolor[gray]{0.93}SARATHI-NMF    & 11.29 & 11.95 & 13.99 & \textbf{16.52} \\
\midrule
\multicolumn{5}{l}{\textit{LLaMA-2-13B} (Dense PPL: 4.88)} \\
SoBP \citep{wei2024sobp} & \textbf{5.38}  & 5.62  & 6.27 & 7.39 \\
\rowcolor[gray]{0.93}SARATHI-Wanda  & --    & 6.36  & 7.12 & 8.25 \\
\rowcolor[gray]{0.93}SARATHI-NMF    & --    & \textbf{5.42} & \textbf{5.85} & \textbf{6.40} \\
\bottomrule
\end{tabular}%
}
\end{table}

\begin{table}[ht]
\caption{Zero-shot accuracy (\%) on LLaMA-2-7B at 30\% sparsity compared to PruneNet~\citep{prunenet2025}. \textbf{Bold} indicates the best result. SARATHI results are detailed in Appendix~\ref{sec:appendix_llama2}.}
\label{tab:prunenet}
\centering
\renewcommand{\arraystretch}{1.1}
\resizebox{\columnwidth}{!}{%
\begin{tabular}{lcccccc}
\toprule
\textbf{Method} & \textbf{PIQA} & \textbf{Hella} & \textbf{Wino} & \textbf{ARC-e} & \textbf{ARC-c} & \textbf{Avg\,(5)} \\
\midrule
PruneNet \citep{prunenet2025}        & 71.11 & \textbf{58.30} & 61.09 & 53.20 & 33.53 & 55.45 \\
\rowcolor[gray]{0.93}SARATHI-NMF     & \textbf{71.33} & 45.62 & 61.48 & \textbf{61.24} & 31.57 & 54.25 \\
\rowcolor[gray]{0.93}SARATHI-Wanda   & \textbf{73.45} & 49.87 & \textbf{66.85} & \textbf{69.70} & \textbf{35.67} & \textbf{59.11} \\
\bottomrule
\end{tabular}%
}
\end{table}

\begin{table*}[t]
\caption{Zero-shot accuracy ($\uparrow$) at 30\% sparsity. \textbf{Bold} indicates the best result. PPL refers to WikiText-2 perplexity. Baselines are sourced from \citet{wei2024sobp} and \citet{dumitru2024change}. $\dagger$ indicates failure due to code errors; $\ddagger$ indicates tasks not reported by the original authors. All SARATHI models are evaluated without fine-tuning.}
\label{tab:main_zs}
\centering
\small
\renewcommand{\arraystretch}{1.1}
\setlength{\tabcolsep}{4pt}
\begin{tabular}{lccccccccc}
\toprule
\textbf{Method} & \textbf{PIQA} & \textbf{Hella} & \textbf{ARC-c} & \textbf{ARC-e} & \textbf{Wino} & \textbf{BoolQ} & \textbf{OBQA} & \textbf{Avg} & \textbf{PPL} \\
\midrule
\multicolumn{10}{l}{\textit{OPT-2.7B}} \\
\rowcolor[gray]{0.97}Dense    & 74.81 & 60.58 & 31.14 & 54.42 & 61.01 & 63.20 & 35.40 & 54.37 & 12.47 \\
FLAP \citep{an2024flap} & 72.14 & 54.82 & 29.27 & 48.86 & 57.46 & 61.28 & 33.20 & 51.00 & 18.32 \\
SliceGPT \citep{ashkboos2024slicegpt} & 72.85 & 55.43 & 30.12 & 49.75 & 58.09 & 59.42 & 33.60 & 51.32 & 16.31 \\
SoBP \citep{wei2024sobp} & 72.69 & \textbf{56.51} & \textbf{30.80} & 50.80 & \textbf{58.88} & 56.09 & \textbf{34.40} & 51.45 & \textbf{13.89} \\
\rowcolor[gray]{0.93}SARATHI-Wanda & 72.96 & 55.94 & 28.67 & \textbf{53.32} & 58.17 & \textbf{65.11} & 33.00 & 52.45 & 15.89 \\
\rowcolor[gray]{0.93}SARATHI-NMF   & \textbf{73.23} & 56.22 & 28.92 & 52.90 & 57.77 & 65.08 & \textbf{34.40} & \textbf{52.65} & 15.61 \\
\midrule
\multicolumn{10}{l}{\textit{OPT-6.7B}} \\
\rowcolor[gray]{0.97}Dense    & 76.22 & 67.20 & 33.50 & 60.48 & 65.11 & 66.35 & 37.40 & 58.04 & 10.86 \\
FLAP \citep{an2024flap} & 74.75 & 62.22 & 32.25 & 56.82 & 62.35 & 61.00 & 36.40 & 55.11 & 15.42 \\
SliceGPT \citep{ashkboos2024slicegpt} & \textbf{75.63} & \textbf{64.20} & \textbf{32.94} & 57.24 & 63.22 & 62.78 & 37.80 & 56.26 & 12.80 \\
SoBP \citep{wei2024sobp} & 75.24 & 63.22 & 32.85 & 57.24 & 61.72 & \textbf{65.29} & 36.60 & 56.02 & \textbf{12.17} \\
\rowcolor[gray]{0.93}SARATHI-Wanda & 75.14 & 62.43 & \textbf{32.94} & 57.41 & \textbf{64.72} & 63.43 & 39.00 & \textbf{56.44} & 14.02 \\
\rowcolor[gray]{0.93}SARATHI-NMF   & 75.46 & 62.46 & 32.00 & \textbf{57.87} & 63.22 & 62.78 & \textbf{39.60} & 56.20 & 13.99 \\
\midrule
\multicolumn{10}{l}{\textit{LLaMA-3-8B}} \\
SliceGPT \citep{ashkboos2024slicegpt} & 59.30 & 37.20 & $\ddagger$ & 42.90 & 56.40 & $\ddagger$ & $\ddagger$ & $\ddagger$ & 13.37 \\
Dyn.~Slicing \citep{dumitru2024change} & 60.40 & 38.40 & $\ddagger$ & 42.40 & 58.00 & $\ddagger$ & $\ddagger$ & $\ddagger$ & 12.96 \\
\rowcolor[gray]{0.93}SARATHI-Wanda & \textbf{71.71} & \textbf{64.94} & \textbf{36.77} & \textbf{63.72} & \textbf{70.72} & \textbf{63.30} & 38.00 & \textbf{58.45} & 9.04 \\
\rowcolor[gray]{0.93}SARATHI-NMF   & 70.08 & 58.08 & 30.89 & 53.66 & 66.61 & 62.54 & \textbf{38.80} & 54.38 & \textbf{8.70} \\
\midrule
\multicolumn{10}{l}{\textit{Mistral-7B}} \\
SliceGPT \citep{ashkboos2024slicegpt} & 62.60 & 38.00 & $\ddagger$ & 51.10 & 59.70 & $\ddagger$ & $\ddagger$ & $\ddagger$ & 8.87 \\
Dyn.~Slicing \citep{dumitru2024change} & 63.10 & 38.60 & $\ddagger$ & 51.70 & 60.20 & $\ddagger$ & $\ddagger$ & $\ddagger$ & 8.76 \\
\rowcolor[gray]{0.93}SARATHI-Wanda & \textbf{66.16} & \textbf{60.92} & \textbf{38.74} & \textbf{65.53} & \textbf{67.40} & \textbf{62.20} & \textbf{35.40} & \textbf{56.62} & 7.74 \\
\rowcolor[gray]{0.93}SARATHI-NMF   & 66.10 & 56.88 & 32.68 & 57.11 & 65.11 & 60.80 & 34.40 & 53.30 & \textbf{7.41} \\
\midrule
\multicolumn{10}{l}{\textit{LLaMA-2-13B}} \\
\rowcolor[gray]{0.97}Dense    & 80.52 & 79.37 & 49.23 & 77.48 & 72.30 & 80.58 & 45.20 & 69.24 & 4.88 \\
SoBP \citep{wei2024sobp} & \textbf{76.50} & \textbf{74.55} & \textbf{47.78} & \textbf{74.45} & \textbf{71.82} & \textbf{79.45} & 43.20 & \textbf{66.82} & 6.27 \\
\rowcolor[gray]{0.93}SARATHI-Wanda & 74.92 & 70.21 & 41.04 & 65.82 & 65.04 & 66.21 & 41.80 & 60.72 & 7.12 \\
\rowcolor[gray]{0.93}SARATHI-NMF   & 74.27 & 69.30 & 40.61 & 67.30 & 70.40 & 72.87 & \textbf{43.60} & 62.62 & \textbf{5.85} \\
\bottomrule
\end{tabular}
\end{table*}

\begin{figure*}[t]
  \centering
  \includegraphics[width=\textwidth]{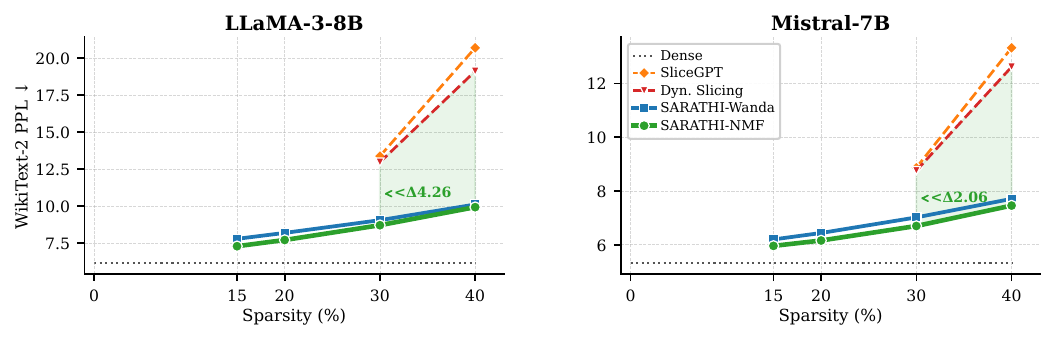}
  \caption{\textbf{WikiText-2 PPL vs.\ sparsity for LLaMA-3-8B (left) and Mistral-7B (right).}
    SARATHI-NMF (green, solid) achieves the lowest PPL at every sparsity level for these two architectures.
    The shaded region between SARATHI-NMF and Dynamic Slicing highlights the growing
    advantage at higher compression rates.
    SliceGPT and Dynamic Slicing results are taken from \citet{dumitru2024change}.
    The dashed horizontal line marks the dense (unpruned) baseline.
    Numbers annotated at 30\% sparsity indicate the absolute PPL gap.}
  \label{fig:ppl_curves}
\end{figure*}

\paragraph{PPL results.}
SARATHI-NMF achieves the lowest PPL on LLaMA-3-8B and Mistral-7B at all tested sparsity levels (Figure~\ref{fig:ppl_curves}), and succeeds at 40\% on OPT-6.7B. On Mistral-7B at 30\%, SARATHI-NMF achieves 7.41 PPL vs.\ 8.76 for Dynamic Slicing ($-15\%$). On LLaMA-2-13B, SARATHI-NMF achieves 5.42 PPL at 20\% and 5.85 PPL at 30\%, outperforming SoBP (5.62 at 20\%, 6.27 at 30\%) without any calibration data for neuron selection. On OPT-2.7B at 30\%, SoBP achieves lower PPL (13.89 vs.\ 15.61 for SARATHI-NMF) through calibration-dependent global reconstruction; however, SoBP fails at 40\% sparsity ($\dagger$), whereas SARATHI-NMF succeeds at 18.98 PPL, and FLAP degrades sharply to 18.32 at 30\%. On OPT-6.7B at 30\% sparsity, SoBP again achieves lower PPL (12.17 vs.\ 13.99 for SARATHI-NMF); this gap arises because SoBP's global reconstruction is more accurate at lower sparsity, but incurs substantial peak GPU memory due to concurrent Hessian inverses, Cholesky factors, and OBS buffers, causing OOM failure at 40\%, the regime where SARATHI-NMF achieves 16.52 PPL.

\paragraph{Zero-shot accuracy.}
On OPT models at 30\% sparsity, SARATHI-NMF achieves higher zero-shot accuracy than SoBP (OPT-2.7B: 52.65\% vs.\ 51.45\%; OPT-6.7B: 56.20\% vs.\ 56.02\%) without any calibration data for neuron selection, despite SoBP attaining lower PPL.
On LLaMA-2-13B, SoBP achieves 66.82\% zero-shot at 30\% (PPL 6.27) by using calibration-dependent global importance selection; SARATHI-NMF reaches 62.62\% (PPL 5.85) without calibration data for selection, and achieves better PPL at both 20\% and 30\% sparsity.
At 40\% on LLaMA-2-13B, SARATHI-NMF obtains 6.40 PPL and 60.98\% zero-shot accuracy, outperforming SoBP on PPL (7.39) while using no calibration data for neuron scoring.
On gated architectures (LLaMA-3-8B, Mistral-7B), SARATHI-Wanda (which uses a 128-sample activation batch for neuron scoring) yields higher zero-shot accuracy than SARATHI-NMF, a trade-off analysed in \S\ref{sec:analysis}: when the calibration domain is well-matched, activation-based scores track task-relevant neuron pathways more tightly. Both SARATHI variants consistently match or exceed SliceGPT and Dynamic Slicing on zero-shot accuracy (see Figure~\ref{fig:radar} for LLaMA-3-8B).

\paragraph{OOM stability.}
At 40\% sparsity on OPT-6.7B, a non-gated ReLU architecture, all coupled baselines encounter OOM failures; SARATHI-Wanda, which retains a 128 calibration batch for weight reconstruction, also fails in this high-sparsity regime.
Among all evaluated methods, SARATHI-NMF successfully complete compression, achieving 16.52 PPL and 54.30\% zero-shot accuracy, consistent with the $\mathcal{O}(d^2)$ peak-memory advantage of the decoupled design.
Extended results at all sparsity levels, LLaMA-2-7B results, and full LLaMA-2-13B breakdowns are in Appendix~\ref{sec:appendix_zs}--\ref{sec:appendix_llama2_13b}.

\paragraph{Comparison with data-free PruneNet.}
We compare against PruneNet~\citep{prunenet2025}, which uses a 45M-parameter
RL policy trained offline to preserve spectral weight structure, making it calibration-free at inference.
PruneNet reports results on LLaMA-2-7B over 5 tasks; Table~\ref{tab:prunenet} uses the same subset.
SARATHI-Wanda outperforms PruneNet by $+3.7$\,pp in average zero-shot accuracy using only
a 128-sample calibration batch and no offline training.
SARATHI-NMF, being fully data-free for selection, achieves comparable aggregate accuracy (54.25\%
vs.\ 55.45\%) while attaining lower perplexity (8.58\,PPL; Table~\ref{tab:llama2}), consistent
with the PPL--accuracy trade-off analysed in \S\ref{sec:analysis}.

\paragraph{Efficiency and Deployment.}
\label{sec:throughput}

Table~\ref{tab:throughput} (Appendix~\ref{sec:appendix_throughput}) reports full per-model throughput and NTK diagnostics. Because SARATHI physically reduces the FFN intermediate dimension, speedup is hardware-agnostic and requires no sparse kernel support. NTK Cosine Similarity is 1.0000 universally, confirming Adaptive OBS preserves local model dynamics across all architectures and sparsity levels. NTK Trace $>$1 for some OPT and LLaMA-2-13B layers reflects bias-correction scaling that amplifies kernel norm while preserving direction.
At BS=16, OPT-6.7B achieves up to $2.05\times$ throughput at 40\% and LLaMA-3-8B achieves $1.66\times$ over the dense baseline. OPT-2.7B gains a modest $1.06$--$1.11\times$ because its smaller FFN is already cache-efficient. Mistral-7B shows a slight throughput regression at high batch sizes because structured removal shifts the FFN intermediate dimension away from Tensor-Core-aligned multiples of 64; at higher sparsity, BS=1 throughput can also slightly decrease as alignment overhead outweighs the reduced compute. Pruning takes 16--44 min on a single A100 with no retraining.

\subsection{Ablation Studies.}
\label{sec:ablation_rank}

\paragraph{NMF rank.}
Table~\ref{tab:ablation} reports WikiText-2 PPL sensitivity to NMF rank $r \in \{7, 16, 32, 64\}$ for LLaMA-3-8B and Mistral-7B at 20--40\% sparsity. PPL varies by at most 0.28 across ranks, validating $r{=}16$ as a robust default.

\begin{table}[ht]
\caption{PPL sensitivity to NMF rank $r$ under Uniform slicing settings. The min-keep floor $\alpha=0.5$ is constant. Best per block \textbf{bolded}.}
\label{tab:ablation}
\centering
\small
\setlength{\tabcolsep}{5pt}
\begin{tabular}{llccc}
\toprule
\textbf{Model} & \textbf{Rank} & \textbf{20\%} & \textbf{30\%} & \textbf{40\%} \\
\midrule
\multirow{4}{*}{LLaMA-3-8B}
 & $r=7$  & 12.38 & 15.78 & 20.17 \\
 & $r=16$ & \textbf{12.37} & \textbf{15.76} & \textbf{20.01} \\
 & $r=32$ & 12.37 & 15.72 & 20.16 \\
 & $r=64$ & 12.39 & 15.73 & 20.02 \\
\midrule
\multirow{4}{*}{Mistral-7B}
 & $r=7$  & 10.75 & 14.37 & \textbf{18.74} \\
 & $r=16$ & 10.86 & 14.39 & 19.01 \\
 & $r=32$ & 10.87 & 14.42 & 19.02 \\
 & $r=64$ & 10.87 & 14.44 & 18.98 \\
\bottomrule
\end{tabular}
\end{table}

\paragraph{Bias Shift Compensation.}
Table~\ref{tab:bias} quantifies the impact. Without bias correction, OPT-2.7B PPL collapses to 2,082 at 40\% sparsity; with correction it drops to 18.98, a 99.1\% reduction vs.\ the uncorrected model.

\begin{table}[ht]
\caption{Effect of Bias Shift Compensation on OPT-2.7B WikiText-2 PPL.}
\label{tab:bias}
\centering
\small
\begin{tabular}{lccc}
\toprule
\textbf{Configuration} & \textbf{20\%} & \textbf{30\%} & \textbf{40\%} \\
\midrule
Without Bias Comp. & 47.06 & 132.00 & 2,082 \\
With Bias Comp. & \textbf{13.25} & \textbf{15.61} & \textbf{18.98} \\
\midrule
PPL Reduction & 72\% & 88\% & \textbf{99.1\%} \\
\bottomrule
\end{tabular}
\end{table}

\paragraph{Adaptive vs.\ Uniform slicing (LLaMA-2-13B).}
Global MAD thresholding adaptively allocates variable FFN widths across layers, preventing bottleneck collapse. Against a rigid uniform baseline (SARATHI-U), the adaptive strategy (SARATHI-A) achieves 5.85 PPL vs.\ 7.48 at 30\%, and 62.62\% vs.\ 60.15\% zero-shot, with NTK Cosine Similarity 1.0000 at every sparsity. Full per-task breakdown is in Appendix~\ref{sec:appendix_llama2_13b}.

\section{Analysis}
\label{sec:analysis}

Two consistent patterns emerge across architectures. First, SARATHI-NMF achieves lower PPL while SARATHI-Wanda achieves higher zero-shot accuracy on gated models (LLaMA-3-8B, Mistral-7B). At 30\% on LLaMA-3-8B, SARATHI-NMF achieves 8.70 PPL ($-3.8\%$ vs.\ Wanda) yet Wanda achieves 58.45\% zero-shot ($+4.1$~pp vs.\ NMF): the NMF probe preserves the generalised language manifold, while the Wanda probe biases the subnet toward commonsense patterns in the calibration corpus.

Second, the NMF probe is most informative for \emph{gated} architectures (SwiGLU/GLU-type FFNs), where gate channels create structured weight-space sparsity that the NMF residual detects. For OPT (ReLU FFNs), scores are near-uniform; Bias Shift Compensation and Adaptive OBS dominate performance, not the probe ranking. NTK Cosine Similarity remains 1.0000 universally, confirming Adaptive OBS fully recovers functional dynamics regardless of probe quality. The NMF vs.\ SVD ablation (Appendix~\ref{sec:appendix_nmf_svd}) further confirms that the non-negativity constraint - not factorization rank - is the decisive factor: a signed SVD residual raises PPL by over 8{,}500 points on OPT-6.7B despite 98.95\% overlap in top-ranked neurons.


\section{Conclusion}
\label{sec:conclusion}

This work presents SARATHI, a structured LLM pruning framework that decouples neuron selection from weight reconstruction.
The data-free NMF Residual Subspace Probe identifies structurally irreplaceable neurons via weight geometry, establishing a formal connection to Robust PCA and providing the first geometric justification for data-free structured pruning.
Adaptive OBS reconstruction operates independently on the kept subnet with $\mathcal{O}(d^2)$ memory per layer, eliminating both calibration-domain bias and the OOM bottleneck of coupled methods.
An empirically discovered Bias Shift Compensation enables stable pruning of post-LayerNorm architectures at sparsity levels where all compared methods fail.
Across six models at 15--40\% sparsity, SARATHI achieves state-of-the-art PPL on gated architectures (LLaMA-3-8B, Mistral-7B), and consistently matches or exceeds existing methods on zero-shot accuracy across all architectures. Future work includes extending the probe to attention heads and non-gated architectures.

\section{Related Work}
\label{sec:related}

\paragraph{Unstructured pruning.}
SparseGPT \citep{frantar2023sparsegpt} extends OBS \citep{hassibi1993obs} to LLMs via approximate inverse-Hessian updates. Wanda \citep{sun2024wanda} scores weights by magnitude times activation norm. FISTAPruner \citep{zhao2025fistapruner} reformulates weight sparsification as a convex LASSO problem solved via FISTA, achieving monotone convergence guarantees. Despite near-lossless compression at 50\% unstructured sparsity, all these methods produce sparse weight tensors that accelerate inference \emph{only} with specialised sparse BLAS kernels (e.g., NVIDIA 2:4 Tensor Cores), limiting their deployment generality.

\paragraph{Structured pruning.}
LLM-Pruner \citep{ma2023llmpruner} uses first-order Taylor scores with LoRA recovery. SliceGPT \citep{ashkboos2024slicegpt} applies PCA-based rotations but injects extra parameters altering layer shapes. FLAP \citep{an2024flap} uses activation variance scores; SoBP \citep{wei2024sobp} introduces a three-stage pipeline (global importance scoring, local greedy refinement, and module-wise OBS reconstruction) using a unified knapsack solver to balance structured pruning budgets across heterogeneous layer types. All four methods construct the selection mask from calibration activations, coupling the mask to $\mathcal{D}_\text{calib}$ and inheriting the domain bias and memory cost analysed in \S\ref{sec:coupling}. PruneNet \citep{prunenet2025} reformulates pruning as a policy-learning process via a 45M-parameter RL learner trained to preserve spectral structure; while calibration-free at inference, it requires a separate offline RL training phase. Table~\ref{tab:prunenet} provides a direct comparison on LLaMA-2-7B.

\paragraph{Matrix factorization for compression.}
SVD-LLM \citep{svdllm2024} and ASVD \citep{yuan2023asvd} truncate singular values to obtain low-rank approximations; LoRA \citep{hu2022lora} constrains fine-tuning updates to low-rank increments. These methods do not physically remove neurons and yield no FLOPs reduction without additional hardware. SARATHI uses NMF as a \emph{geometric probe}, not to approximate weights, but to identify and physically remove redundant neurons, yielding genuinely smaller dense matrices. NMF has previously been explored for general DNN compression \citep{gurve2024on} and data-free pruning at initialization \citep{behera2026you}; however, those works operate in fundamentally different settings (pre-LLM architectures, at-initialization pruning) and do not address post-training LLM structured pruning. SARATHI is distinct: it applies NMF as a \emph{residual subspace probe} for post-training LLM pruning, with a formal Robust PCA safety bound (Proposition~\ref{prop:nmf}), MAD-based global budget allocation, and a decoupled Adaptive OBS reconstruction and bias-correction pipeline that are absent from prior NMF-based compression methods.

\section*{Limitations}

\paragraph{NMF probe on non-gated architectures.}
For ReLU-FFN models such as OPT, NMF scores are near-uniform and offer limited discriminative signal. SARATHI still achieves strong PPL on these models because Bias Shift Compensation and Adaptive OBS are the dominant contributors; the NMF probe's benefit is concentrated in gated architectures. Activation-conditioned probing for non-gated models is a natural direction for future work.

\paragraph{Scope.}
PPL evaluation uses WikiText-2 only. Throughput gains at 15--20\% sparsity are modest, below 10\%, due to cache-line alignment effects. On Mistral-7B, structured removal shifts FFN dimensions away from Tensor-Core-aligned sizes, which introduces a throughput regression of 0.82--0.92$\times$ relative to the dense baseline. NMF selection adds roughly 5--10\% overhead over magnitude-only scoring but requires no gradient computation. On LLaMA-2-13B at 30\%, SoBP's calibration-dependent mask achieves higher zero-shot accuracy, 66.82\% vs.\ 62.62\% for SARATHI-NMF, though SARATHI-NMF attains lower PPL, 5.85 vs.\ 6.27. This trade-off reflects the broader property that data-free probing favours perplexity fidelity and memory safety over calibration-domain task alignment.

\section*{Ethical Considerations}

The models and datasets used in this work are publicly available and do not contain personally identifiable information.
Structured pruning as studied here reduces the computational cost of LLM inference, potentially reducing energy consumption for deployed systems.
No new training data is introduced.

\section*{Acknowledgments}
The authors would like to thank Dr.\ Raj Kumar Choudhary, Engineering College Bikaner, for providing
computational resources and access to the computing facilities under the ISEA Phase-III project, which
significantly supported this research.

\bibliography{sarathi_refs}

\appendix

\section{Throughput, Pruning Time, and NTK Diagnostics}
\label{sec:appendix_throughput}

Table~\ref{tab:throughput} reports inference throughput (tokens/sec on A100 40GB), speedup over the dense baseline, pruning time, and NTK diagnostics across five of the evaluated models and sparsity levels. Because SARATHI physically reduces the FFN intermediate dimension, speedup is hardware-agnostic and requires no sparse kernel support. Pruning takes 16--44 min on a single A100 40GB GPU with no retraining.

\begin{table*}[h]
\caption{Hardware performance and structural diagnostics for SARATHI-NMF. Throughput is measured on a single A100 40GB GPU. NTK Cosine Similarity near 1.0000 indicates preserved local gradient dynamics. Mistral-7B exhibits a modest throughput regression at BS=16 due to FFN dimensions shifting away from Tensor-Core-aligned sizes.}
\label{tab:throughput}
\centering
\small
\setlength{\tabcolsep}{3.5pt}
\begin{tabular}{llcccccc}
\toprule
\textbf{Model} & \textbf{Sparsity} & \textbf{BS=1} & \textbf{BS=16} & \textbf{Speedup} & \textbf{NTK Cos.} & \textbf{NTK Trace} & \textbf{Prune Time} \\
\midrule
\multirow{4}{*}{OPT-2.7B}
 & Dense (0\%) & 46.52 & 699.30 & 1.00$\times$ & -- & -- & -- \\
 & 20\% & 51.28 & 779.17 & 1.11$\times$ & 1.0000 & 0.9639 & 16.5 min \\
 & 30\% & 46.55 & 759.22 & 1.09$\times$ & 1.0000 & 1.0495 & 17.7 min \\
 & 40\% & 48.10 & 740.95 & 1.06$\times$ & 1.0000 & 1.1277 & 18.4 min \\
\midrule
\multirow{4}{*}{OPT-6.7B}
 & Dense (0\%) & 44.12 & 358.39 & 1.00$\times$ & -- & -- & -- \\
 & 20\% & 53.05 & 646.64 & 1.80$\times$ & 1.0000 & 1.5222 & 33.7 min \\
 & 30\% & 52.37 & 709.37 & 1.98$\times$ & 1.0000 & 1.1124 & 34.1 min \\
 & 40\% & 52.95 & 734.14 & 2.05$\times$ & 1.0000 & 1.2157 & 36.0 min \\
\midrule
\multirow{4}{*}{LLaMA-3-8B}
 & Dense (0\%) & 30.83 & 283.88 & 1.00$\times$ & -- & -- & -- \\
 & 20\% & 30.99 & 471.67 & 1.66$\times$ & 1.0000 & 0.4390 & 28.8 min \\
 & 30\% & 31.21 & 472.43 & 1.66$\times$ & 1.0000 & 0.2598 & 31.5 min \\
 & 40\% & 30.03 & 457.11 & 1.61$\times$ & 1.0000 & 0.1678 & 33.1 min \\
\midrule
\multirow{4}{*}{Mistral-7B}
 & Dense (0\%) & 34.84 & 542.33 & 1.00$\times$ & -- & -- & -- \\
 & 20\% & 35.61 & 499.26 & 0.92$\times$ & 1.0000 & 0.2654 & 33.2 min \\
 & 30\% & 35.08 & 486.04 & 0.90$\times$ & 1.0000 & 0.1855 & 35.2 min \\
 & 40\% & 30.31 & 444.29 & 0.82$\times$ & 1.0000 & 0.1176 & 34.8 min \\
\midrule
\multirow{4}{*}{LLaMA-2-13B}
 & Dense (0\%) & 14.20 & -- & 1.00$\times$ & -- & -- & -- \\
 & 20\% & 18.77 & -- & 1.32$\times$ & 1.0000 & 2.0855 & 38.9 min \\
 & 30\% & 18.64 & -- & 1.31$\times$ & 1.0000 & 0.3186 & 42.6 min \\
 & 40\% & 18.23 & -- & 1.28$\times$ & 1.0000 & 0.3307 & 43.7 min \\
\bottomrule
\end{tabular}
\end{table*}


\section{Proof Sketch for Proposition~\ref{prop:nmf}}
\label{sec:proof_prop1}

\paragraph{Setup.}
Let $(\mathbf{V}^*, \mathbf{H}^*)$ be a stationary point of the NMF objective (Eq.~\ref{eq:nmf}), so that the rank-$r$ approximation satisfies $|\mathbf{W}^{(\ell)}| \approx \mathbf{V}^*\mathbf{H}^*$ with residual $\mathbf{S} = |\mathbf{W}^{(\ell)}| - \mathbf{V}^*\mathbf{H}^*$.
The element-wise binary mask is $\mathcal{M}_{i,j} = \mathbb{I}(\mathbf{S}_{i,j} \geq \tau)$, and the importance score is $\eta_i = \sum_j (|\mathbf{W}^{(\ell)}_{i,j}| \cdot \mathcal{M}_{i,j})$.

\paragraph{Part 1: Safe removal of zero-$\eta$ neurons.}
Suppose $\eta_i = 0$, i.e., $\mathbf{S}_{i,j} < \tau$ for all $j$.
This implies $\|\mathbf{S}_{i,:}\|_\infty < \tau$, and therefore $\|\mathbf{S}_{i,:}\|_2 < \tau\sqrt{d}$:
\begin{equation*}
  |\mathbf{W}^{(\ell)}_{i,:}| = (\mathbf{V}^*\mathbf{H}^*)_{i,:} + \mathbf{S}_{i,:},
  \quad \|\mathbf{S}_{i,:}\|_2 < \tau\sqrt{d}.
\end{equation*}
The layer output contribution of neuron $i$ over a token batch $\mathbf{X}$ is $\mathbf{W}^{(\ell)}_{i,:}\mathbf{X} \in \mathbb{R}^T$.
Removing row $i$ and redistributing its contribution via the Adaptive OBS weight update (Eq.~\ref{eq:obs}) yields a residual reconstruction error bounded by:
\begin{equation*}
  \|\mathbf{W}^{(\ell)}_{i,:}\mathbf{X} - \hat{\mathbf{W}}_{i,:}\mathbf{X}\|_2
  \leq \|\mathbf{S}_{i,:}\|_2 \cdot \|\mathbf{X}\|_F
  < \tau\sqrt{d}\,\|\mathbf{X}\|_F,
\end{equation*}
where $\hat{\mathbf{W}}_{i,:}$ is the post-Adaptive OBS approximation.
The total output reconstruction error for the full layer after removing neuron $i$ is thus $\mathcal{O}(\tau\sqrt{d}\,\|\mathbf{X}\|_F)$.

\paragraph{Part 2: Irreplaceability of nonzero-$\eta$ neurons.}
Conversely, suppose $\eta_j > 0$, i.e., there exists at least one $j'$ with $\mathbf{S}_{j,j'} \geq \tau$, meaning $|\mathbf{W}^{(\ell)}_{j,j'}|$ lies outside the NMF consensus.
By definition, $(\mathbf{V}^*\mathbf{H}^*)_{j,:}$ is the best rank-$r$ NMF approximation to $|\mathbf{W}^{(\ell)}_{j,:}|$ achievable within $\mathrm{col}(\mathbf{V}^{*\top})$.
If $\|\mathbf{S}_{j,:}\|_2 = \delta > 0$, then $|\mathbf{W}^{(\ell)}_{j,:}|$ lies at distance $\delta$ from the subspace $\mathrm{col}(\mathbf{V}^{*\top})$: it cannot be expressed as a non-negative linear combination of the basis vectors $\{\mathbf{H}^*_{k,:}\}_{k=1}^r$ with error less than $\delta$.
Therefore, no convex combination of the remaining $N-1$ rows (all of which lie in or near $\mathrm{col}(\mathbf{V}^{*\top})$) can reconstruct the output of neuron $j$, making its removal irreversible within the low-rank subspace.

\paragraph{Remark on stationarity.}
The multiplicative update rules for NMF (Eq.~\ref{eq:nmf_update}) are guaranteed to converge to a stationary point of the Frobenius-norm objective~\citep{lee1999nmf}, though not necessarily a global minimum.
The bounds above hold at any stationary point; tighter bounds at global minima follow directly by substituting the globally optimal residual $\mathbf{S}^*$ in place of $\mathbf{S}$.

\section{Extended Zero-Shot Accuracy Results}
\label{sec:appendix_zs}

Table~\ref{tab:zs_extended} reports zero-shot accuracy at 20\% and 40\% sparsity for all four evaluated models, supplementing the 30\% results in the main text (Table~\ref{tab:main_zs}).
At 40\% sparsity on OPT-6.7B, SARATHI-Wanda (vB) runs out of memory; only SARATHI-NMF results are reported.

\begin{table*}[h]
\caption{Extended zero-shot accuracy ($\uparrow$) at 20\% and 40\% sparsity. SARATHI-W = SARATHI-Wanda; SARATHI-N = SARATHI-NMF. OOM = out-of-memory.}
\label{tab:zs_extended}
\centering
\small
\setlength{\tabcolsep}{3pt}
\begin{tabular}{llccccccccc}
\toprule
\textbf{Model} & \textbf{Method} & \textbf{Spar.} & \textbf{PIQA} & \textbf{Hella} & \textbf{ARC-c} & \textbf{ARC-e} & \textbf{Wino} & \textbf{BoolQ} & \textbf{OBQA} & \textbf{Avg} \\
\midrule
\multirow{4}{*}{OPT-2.7B}
 & SARATHI-W & 20\% & 73.88 & 59.02 & 29.69 & 53.24 & 60.06 & 65.57 & 35.60 & 53.87 \\
 & SARATHI-N & 20\% & 73.99 & 59.19 & 29.10 & 54.00 & 59.12 & 63.21 & 35.00 & 53.37 \\
 & SARATHI-W & 40\% & 71.38 & 51.83 & 28.50 & 49.92 & 58.72 & 64.19 & 32.00 & 50.93 \\
 & SARATHI-N & 40\% & 71.87 & 51.98 & 28.41 & 51.01 & 58.25 & 64.04 & 31.80 & 51.05 \\
\midrule
\multirow{3}{*}{OPT-6.7B}
 & SARATHI-W & 20\% & 76.44 & 65.28 & 32.94 & 59.47 & 65.04 & 67.25 & 38.40 & 57.83 \\
 & SARATHI-N & 20\% & 76.33 & 65.32 & 32.68 & 60.14 & 65.19 & 66.64 & 38.60 & 57.84 \\
 & SARATHI-N & 40\% & 74.05 & 59.15 & 30.12 & 54.50 & 62.43 & 62.23 & 37.60 & 54.30 \\
\midrule
\multirow{4}{*}{LLaMA-3-8B}
 & SARATHI-W & 20\% & 74.81 & 70.86 & 43.34 & 68.35 & 71.43 & 63.49 & 42.20 & 62.07 \\
 & SARATHI-N & 20\% & 73.39 & 63.84 & 32.25 & 58.46 & 68.51 & 63.30 & 41.00 & 57.25 \\
 & SARATHI-W & 40\% & 66.76 & 57.34 & 32.59 & 53.75 & 69.14 & 63.03 & 36.40 & 54.14 \\
 & SARATHI-N & 40\% & 68.01 & 52.93 & 29.78 & 48.19 & 64.96 & 62.23 & 35.60 & 51.67 \\
\midrule
\multirow{4}{*}{Mistral-7B}
 & SARATHI-W & 20\% & 71.82 & 70.04 & 44.71 & 71.46 & 71.19 & 80.46 & 39.20 & 64.13 \\
 & SARATHI-N & 20\% & 70.89 & 65.27 & 37.63 & 64.60 & 68.35 & 62.94 & 39.20 & 58.41 \\
 & SARATHI-W & 40\% & 61.70 & 51.87 & 30.89 & 54.04 & 62.83 & 62.02 & 31.80 & 50.74 \\
 & SARATHI-N & 40\% & 62.62 & 49.72 & 29.86 & 52.02 & 62.04 & 58.99 & 31.00 & 49.46 \\
\midrule
\multirow{4}{*}{LLaMA-2-13B}
 & SARATHI-W & 20\% & 77.20 & 73.69 & 42.58 & 67.76 & 66.69 & 69.48 & 41.20 & 62.60 \\
 & SARATHI-N & 20\% & 77.20 & 72.86 & 43.34 & 69.95 & 71.27 & 73.55 & 43.00 & 64.45 \\
 & SARATHI-W & 40\% & 72.52 & 65.45 & 37.71 & 58.84 & 64.09 & 60.98 & 38.20 & 56.83 \\
 & SARATHI-N & 40\% & 71.98 & 65.82 & 37.54 & 63.51 & 70.24 & 75.54 & 42.20 & 60.98 \\
\bottomrule
\end{tabular}
\end{table*}

\section{LLaMA-2-7B Results}
\label{sec:appendix_llama2}

Table~\ref{tab:llama2} reports SARATHI results on LLaMA-2-7B \citep{touvron2023llama2} at 15--40\% structured sparsity.
On this architecture, SARATHI-Wanda (activation-based probe) outperforms SARATHI-NMF on both PPL and zero-shot accuracy. This aligns with our analysis in \S\ref{sec:analysis}: LLaMA-2 exhibits different calibration-domain sensitivity than LLaMA-3, allowing the activation-based probe to better track its commonsense reasoning pathways.

\begin{table*}[h]
\caption{WikiText-2 PPL and average zero-shot accuracy on LLaMA-2-7B. SARATHI-W = SARATHI-Wanda; SARATHI-N = SARATHI-NMF.}
\label{tab:llama2}
\centering
\small
\setlength{\tabcolsep}{4pt}
\begin{tabular}{lcccccccccc}
\toprule
\textbf{Method} & \textbf{Spar.} & \textbf{PPL} & \textbf{PIQA} & \textbf{Hella} & \textbf{ARC-c} & \textbf{ARC-e} & \textbf{Wino} & \textbf{BoolQ} & \textbf{OBQA} & \textbf{Avg ZS} \\
\midrule
Dense & 0\% & 5.47 & -- & -- & -- & -- & -- & -- & -- & -- \\
\midrule
SARATHI-W & 15\% & \textbf{5.83} & 76.44 & 54.76 & 40.44 & 73.36 & 67.72 & 77.00 & 31.20 & \textbf{60.13} \\
SARATHI-N & 15\% & 6.76 & 74.97 & 51.16 & 37.54 & 68.39 & 65.35 & 66.02 & 29.00 & 56.06 \\
\midrule
SARATHI-W & 20\% & \textbf{6.02} & 75.84 & 53.37 & 38.99 & 72.73 & 67.80 & 75.32 & 31.40 & \textbf{59.35} \\
SARATHI-N & 20\% & 7.26 & 74.70 & 49.47 & 34.98 & 66.62 & 62.59 & 62.78 & 26.00 & 53.88 \\
\midrule
SARATHI-W & 30\% & \textbf{6.54} & 73.45 & 49.87 & 35.67 & 69.70 & 66.85 & 67.55 & 27.20 & \textbf{55.76} \\
SARATHI-N & 30\% & 8.58 & 71.33 & 45.62 & 31.57 & 61.24 & 61.48 & 56.67 & 25.60 & 50.50 \\
\midrule
SARATHI-W & 40\% & \textbf{7.36} & 69.64 & 45.43 & 31.91 & 64.69 & 63.93 & 67.86 & 25.20 & \textbf{52.67} \\
SARATHI-N & 40\% & 10.20 & 68.77 & 42.28 & 30.38 & 55.26 & 60.14 & 49.24 & 22.20 & 46.90 \\
\bottomrule
\end{tabular}
\end{table*}

\section{LLaMA-2-13B Full Results}
\label{sec:appendix_llama2_13b}

Table~\ref{tab:llama2_13b_full} reports per-task zero-shot accuracy and WikiText-2 PPL for LLaMA-2-13B across 20--40\% structured sparsity, comparing a uniform slicing baseline (SARATHI-U) against our adaptive global allocation strategy (SARATHI-A).
The adaptive strategy substantially outperforms the uniform baseline, particularly on Winogrande (+5.14 to +6.67 pp) and BoolQ (+4.68 to +15.23 pp), where the layer-wise bottleneck collapse under uniform pruning is most pronounced.


\begin{table*}[h]
\caption{Per-task zero-shot accuracy (\%) and WikiText-2 PPL for LLaMA-2-13B. SARATHI-U = Uniform + OBS; SARATHI-A = Adaptive + OBS (our method). Best results \textbf{bolded}. Dense baseline avg.\ = 69.24\%.}
\label{tab:llama2_13b_full}
\centering
\small
\setlength{\tabcolsep}{3pt}
\begin{tabular}{llccccccccc}
\toprule
\textbf{Sparsity} & \textbf{Method} & \textbf{PPL} & \textbf{PIQA} & \textbf{Hella} & \textbf{ARC-c} & \textbf{ARC-e} & \textbf{Wino} & \textbf{BoolQ} & \textbf{OBQA} & \textbf{Avg} \\
\midrule
0\% & Dense & 4.88 & 80.52 & 79.37 & 49.23 & 77.48 & 72.30 & 80.58 & 45.20 & 69.24 \\
\midrule
\multirow{2}{*}{20\%}
 & SARATHI-U & 6.86 & 76.21 & 73.09 & 42.06 & 67.19 & 66.13 & 68.87 & 40.62 & 62.02 \\
 & SARATHI-A & \textbf{5.42} & \textbf{77.20} & 72.86 & \textbf{43.34} & \textbf{69.95} & \textbf{71.27} & \textbf{73.55} & \textbf{43.00} & \textbf{64.45} \\
\midrule
\multirow{2}{*}{30\%}
 & SARATHI-U & 7.48 & \textbf{74.32} & \textbf{69.63} & 40.57 & 65.18 & 64.51 & 65.62 & 41.21 & 60.15 \\
 & SARATHI-A & \textbf{5.85} & 74.27 & 69.30 & \textbf{40.61} & \textbf{67.30} & \textbf{70.40} & \textbf{72.87} & \textbf{43.60} & \textbf{62.62} \\
\midrule
\multirow{2}{*}{40\%}
 & SARATHI-U & 8.55 & 71.87 & 64.82 & 37.02 & 58.14 & 63.57 & 60.31 & 37.57 & 56.19 \\
 & SARATHI-A & \textbf{6.40} & 71.98 & \textbf{65.82} & 37.54 & \textbf{63.51} & \textbf{70.24} & \textbf{75.54} & \textbf{42.20} & \textbf{60.98} \\
\bottomrule
\end{tabular}
\end{table*}

The NTK Cosine Similarity remains 1.0000 at all sparsities for LLaMA-2-13B (Table~\ref{tab:throughput}), confirming that the Adaptive OBS reconstruction preserves exact gradient geometry even at 40\% parameter removal from a 13B model.
Pruning LLaMA-2-13B requires 38.9--43.7 minutes on a single A100 40GB GPU without calibration data.

\section{NMF vs.\ SVD Probe Ablation}
\label{sec:appendix_nmf_svd}

To isolate the effect of the factorization type in the SARATHI probe, we replace the NMF decomposition
(Eq.~\ref{eq:nmf}) with a rank-16 truncated SVD while keeping the residual-based importance scoring
(Eq.~\ref{eq:score}) identical. Critically, \emph{Adaptive OBS reconstruction and Bias Shift Compensation
are intentionally excluded} so that any performance difference reflects the probe alone rather than
downstream correction steps.

Table~\ref{tab:nmf_svd_ablation} reports the results of 30\% structured pruning on OPT-6.7B.

\begin{table}[H]
\caption{NMF vs.\ SVD probe ablation on OPT-6.7B at 30\% sparsity (probe-only;
Adaptive OBS and Bias Shift Compensation excluded to isolate probe quality).
Absolute PPL values are not representative of the full SARATHI pipeline; see Table~\ref{tab:main_ppl}.
\textbf{Bold} = better per metric.}
\label{tab:nmf_svd_ablation}
\centering
\small
\resizebox{\columnwidth}{!}{%
\begin{tabular}{lcc}
\toprule
\textbf{Metric} & \textbf{NMF Probe} & \textbf{SVD Probe} \\
\midrule
\multicolumn{3}{l}{\textit{Ranking agreement (probe-independent)}} \\
\quad Spearman $\rho$ (neuron ranking) & 0.9987 & 0.9987 \\
\quad Top-25\% neuron overlap         & 98.95\% & 98.95\% \\
\midrule
\multicolumn{3}{l}{\textit{Reconstruction and downstream quality}} \\
\quad Rel.\ reconstruction error       & 0.6103 & \textbf{0.6078} \\
\quad Coefficient of variation (CV)    & \textbf{0.2805} & 0.2777 \\
\quad WikiText-2 PPL (probe-only)      & \textbf{19{,}695} & 28{,}268 \\
\bottomrule
\end{tabular}%
}
\end{table}

\paragraph{Key finding.}
Despite a Spearman correlation of $\rho = 0.9987$ and a 98.95\% overlap in the top-25\% most important
neurons, replacing NMF with an SVD-based residual raises perplexity by more than 8{,}500 points.
This result demonstrates that \emph{non-negativity in the factorization} - not the rank or the
residual-scoring rule - is the decisive factor. The non-negative additive decomposition enforces
a parts-based structure that aligns with the non-negative activation regime of SwiGLU/ReLU FFNs,
capturing structurally irreplaceable neurons that a signed SVD factorization fails to identify.

\paragraph{Clarification on absolute PPL.}
The perplexity values in Table~\ref{tab:nmf_svd_ablation} are intentionally high because Adaptive OBS
reconstruction and Bias Shift Compensation (\S\ref{sec:obs}) are excluded.
For reference, the full SARATHI-NMF pipeline achieves 11.95, 13.99, and 16.52 PPL on OPT-6.7B at
20\%, 30\%, and 40\% sparsity respectively (Table~\ref{tab:main_ppl}).

\section{Why FFN? Attention vs.\ FFN Parameter Budget in Modern LLMs}
\label{sec:appendix_ffn_motivation}

Structured pruning of LLMs can target different architectural components: attention heads, FFN intermediate neurons, or entire layers. This work focuses on FFN intermediate neurons, motivated by three architectural properties that make FFN layers the dominant target for structured compression. In every modern LLM evaluated here, the FFN sub-layer contains substantially more parameters per transformer layer than the attention sub-layer, and this gap has widened considerably with the adoption of GQA \citep{ainslie2023gqa}.

Table~\ref{tab:ffn_vs_attn} quantifies the per-layer parameter counts for the attention
projection matrices
($\mathbf{W}_Q, \mathbf{W}_K, \mathbf{W}_V, \mathbf{W}_O \in \mathbb{R}^{d \times d}$ for MHA;
$\mathbf{W}_Q \in \mathbb{R}^{d \times d}$, $\mathbf{W}_K, \mathbf{W}_V \in \mathbb{R}^{d \times d/g}$ for GQA with $g$ KV groups)
and the FFN projection matrices
($\mathbf{W}_\text{up}, \mathbf{W}_\text{down} \in \mathbb{R}^{d_\text{ff} \times d}$,
plus $\mathbf{W}_\text{gate}$ for SwiGLU, counted as $3 d \cdot d_\text{ff}$ total).

\begin{table}[h]
\caption{Per-layer parameter counts and Attn\,:\,FFN ratio for representative LLMs.
Attn\,/\,layer counts $\mathbf{W}_Q, \mathbf{W}_K, \mathbf{W}_V, \mathbf{W}_O$.
FFN\,/\,layer counts up, down (and gate for SwiGLU).
GQA models use fewer KV heads (column ``KV\,/\,Q heads''), sharply
reducing attention parameters while leaving the FFN unchanged.}
\label{tab:ffn_vs_attn}
\centering
\small
\setlength{\tabcolsep}{4pt}
\resizebox{\columnwidth}{!}{%
\begin{tabular}{lcccccc}
\toprule
\textbf{Model} & \textbf{Attn} & \textbf{$d$} & \textbf{$d_{ff}$} & \textbf{KV\,/\,Q} & \textbf{Attn\,/\,layer} & \textbf{FFN\,/\,layer} \\
               &               &              &                   & \textbf{heads}    & (M params)              & (M params) \\
\midrule
OPT-66B      & MHA & 9216 & 36\,864 & 72\,/\,72 & 339.74 & 679.48 \\
LLaMA-2-7B   & MHA & 4096 & 11\,008 & 32\,/\,32 &  67.11 & 135.27 \\
\midrule
Mistral-7B   & GQA & 4096 & 14\,336 &  8\,/\,32 &  41.94 & 176.16 \\
LLaMA-3-8B   & GQA & 4096 & 14\,336 &  8\,/\,32 &  41.94 & 176.16 \\
LLaMA-2-70B  & GQA & 8192 & 28\,672 &  8\,/\,64 & 150.99 & 704.64 \\
LLaMA-3-70B  & GQA & 8192 & 28\,672 &  8\,/\,64 & 150.99 & 704.64 \\
\midrule
\multicolumn{5}{l}{\textit{Attn\,:\,FFN ratio}} & & \\
\multicolumn{5}{l}{\quad MHA models (OPT-66B, LLaMA-2-7B)} & \multicolumn{2}{c}{$\approx 1 : 2.0$} \\
\multicolumn{5}{l}{\quad GQA models (Mistral-7B, LLaMA-3-8B)} & \multicolumn{2}{c}{$\approx 1 : 4.2$} \\
\multicolumn{5}{l}{\quad GQA models (LLaMA-2-70B, LLaMA-3-70B)} & \multicolumn{2}{c}{$\approx 1 : 4.7$} \\
\bottomrule
\end{tabular}%
}
\end{table}

Even in older MHA architectures such as OPT-66B, the FFN contains twice as many parameters
as the attention sub-layer.
For GQA-based models which include every state-of-the-art LLM evaluated in
this work (Mistral-7B, LLaMA-3-8B) and large-scale counterparts (LLaMA-2-70B, LLaMA-3-70B), the
FFN accounts for over 80\% of per-layer parameters.
Pruning attention in these architectures yields at most a $\sim$17\% per-layer reduction even
if all attention parameters were removed, whereas a 30\% FFN sparsity directly eliminates 30\%
of the dominant parameter block.

Beyond raw parameter counts, FFN layers have additional structural advantages for compression.
In LLaMA-2-7B, FFN layers collectively contribute 64\% of parameters while self-attention
contributes 32\%; crucially, the neuron density (number of neurons directly connected to a parameter)
in self-attention is only $\sim$27\% due to the multi-headed structure, whereas it is 100\% in FFN
layers \citep{prunenet2025}.
This means each removed FFN neuron eliminates a fully dense row, yielding maximum effective sparsity
per removed parameter.
Additionally, non-linear activations are exclusive to FFN layers (SwiGLU, ReLU), and \citet{mirzadeh2023relu}
highlighted that these non-linear operations comprise a significant fraction of per-layer compute;
pruning FFN neurons thus simultaneously reduces both parameter count and activation overhead.

This strongly motivates the design choice to focus structured pruning on the FFN intermediate
dimension, where the compression leverage is largest, consistent with PruneNet \citep{prunenet2025},
FLAP \citep{an2024flap}, and SoBP \citep{wei2024sobp}.
Extending the NMF Residual Probe to attention heads (e.g., structured head or head-group pruning)
is a natural direction for future work.

\end{document}